\documentclass[letterpaper, 10 pt, conference]{ieeeconf}  

\IEEEoverridecommandlockouts                              
                                                          
\usepackage{tabularx}
\usepackage{amsmath}
\usepackage{subcaption}
\usepackage{graphicx}
\usepackage{multirow}
\usepackage{array}
\usepackage{booktabs}
\usepackage{hyperref}
\usepackage{adjustbox}
\let\labelindent\relax
\usepackage{amssymb}
\usepackage{xcolor}
\usepackage[normalem]{ulem}
\usepackage{pifont}
\usepackage{dcolumn}
\usepackage{enumitem}
\usepackage{setspace}
\setlist{leftmargin=7pt}
\newcolumntype{A}{D{.}{\textrm{\ding{51}}}{-1}}

\usepackage{ntheorem}
\theoremstyle{nonumberplain}

\usepackage{sidecap}

\newcolumntype{Y}{>{\centering\arraybackslash}X}

\def\code#1{\texttt{#1}}

\begin{document}
\title{Personalized Robotic Object Rearrangement from Scene Context}
\author{
    Kartik Ramachandruni and Sonia Chernova
    \thanks{
        Georgia Institute of Technology, Atlanta, Georgia, United States.
        Contact: \texttt{\{kvr,chernova\}@gatech.edu}
    }
}

\maketitle

\begin{abstract}
Object rearrangement is a key task for household robots requiring personalization without explicit instructions, meaningful object placement in environments occupied with objects, and generalization to unseen objects and new environments. 
To facilitate research addressing these challenges, we introduce PARSEC, an object rearrangement benchmark for learning user organizational preferences from observed scene context to place objects in a partially arranged environment. PARSEC is built upon a novel dataset of 110K rearrangement examples crowdsourced from 72 users, featuring 93 object categories and 15 environments.
To better align with real-world organizational habits, we propose ContextSortLM, an LLM-based personalized rearrangement model that handles flexible user preferences by explicitly accounting for objects with multiple valid placement locations when placing items in partially arranged environments.
We evaluate ContextSortLM and existing personalized rearrangement approaches on the PARSEC benchmark and complement these findings with a crowdsourced evaluation of 108 online raters ranking model predictions based on alignment with user preferences.
Our results indicate that personalized rearrangement models leveraging multiple scene context sources perform better than models relying on a single context source. Moreover, ContextSortLM outperforms other models in placing objects to replicate the target user's arrangement and ranks among the top two in all three environment categories, as rated by online evaluators. Importantly, our evaluation highlights challenges associated with modeling environment semantics across different environment categories and provides recommendations for future work.
\end{abstract}

\begin{SCfigure*}
    \centering
    \includegraphics[width=0.7\textwidth]{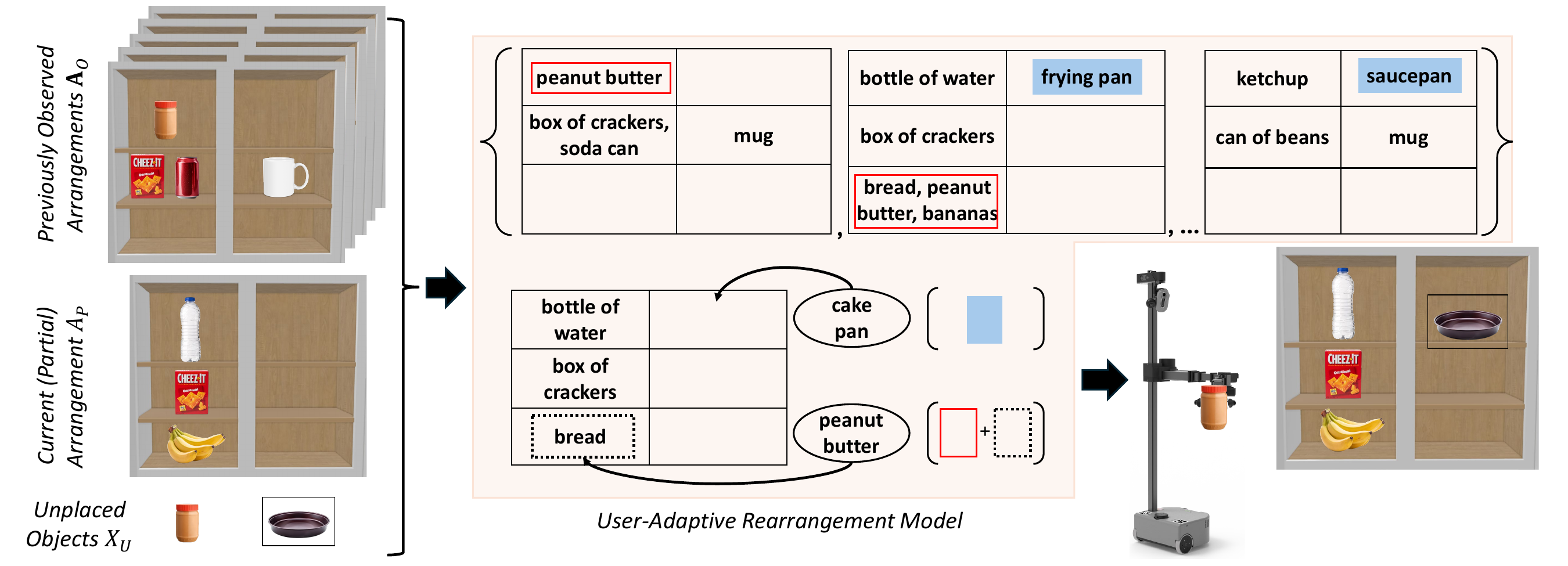}
    \caption{\small{In the PARSEC benchmark, the robot adapts to a user's organizational preferences when placing new objects $\mathcal{X}_U$ in an partially arranged, or pre-occupied, environment. The robot must jointly reason across prior observations of the user arranging objects, $\mathbf{A}_O$, and the environment's current object arrangement, $\mathcal{A}_P$, to meaningfully place objects.}}
    \label{fig:problem-formulation}
    \vspace{-1.4em}
\end{SCfigure*}

\vspace{-0.5em}
\section{Introduction}
Consider a robot that assists users with tidying the home by putting away objects, or what is known as the rearrangement problem~\cite{batra2020rearrangement}. How should this robot determine the appropriate location to put each object?
The robot must develop an object placement strategy based on a user's organizational preferences without detailed instructions or demonstrations, as doing so would unnecessarily burden that user and would need to be repeated for new objects or environments. Furthermore, the robot should ensure that its actions align with the current arrangement of objects in the home (e.g., arrange new pantry items in accordance with existing ones.)  Critically, these capabilities must apply to previously unseen household environments.
Thus, personalized object rearrangement presents three practical challenges: inferring user preferences without explicit instruction to determine the desired goal, meaningfully placing objects in pre-occupied environments, and adapting to new environments.

Prior work has proposed various approaches to infer rearrangement preferences without explicit instruction~\cite{newman2024degustabot, trabucco2022simple, abdo2016organizing, ramrakhya2024seeing, ramachandruni2023consor, brawner2016learning, wang2024apricot}.
Most methods rely on either \textit{prior} scene context~\cite{newman2024degustabot, trabucco2022simple, abdo2016organizing} derived from prior observations of the user arranging objects, and \textit{current} scene context~\cite{ramrakhya2024seeing, ramachandruni2023consor} from the positions of objects already placed in partially arranged environments, to learn user preferences.
However, observation-based rearrangement models assume that the current environment is empty, and \textit{current} scene based models do not perform well in environments sparsely occupied with objects.
Few methods combined both context sources~\cite{brawner2016learning, wang2024apricot}, but either require additional user interaction to infer preferences~\cite{wang2024apricot} or fail to handle unseen objects and new environments~\cite{brawner2016learning}.
Overall, approaches in prior work do not address all the aforementioned challenges.

To facilitate more research in personalized rearrangement, we present PARSEC, an object tidying benchmark and dataset addressing user personalization, object placement in partially arranged environments, and generalization to unseen objects and new environments. In this benchmark, a robot learns user preferences by leveraging \textit{prior} and \textit{current} scene context to determine object placements in environments occupied with objects.
Existing datasets fall short of our objective, either due to lack of real-user data~\cite{wu2023tidybot, ramachandruni2023consor}, omission of user preferences~\cite{kant2022housekeep}, or limited objects and environments~\cite{toris2015unsupervised, kapelyukh2022my}. To address this, we collect a novel crowdsourced dataset of real users arranging various household objects to complete organizational tasks, such as stocking the kitchen pantry and organizing the fridge. 
We evaluate existing rearrangement approaches across different environment types and initial conditions in PARSEC and complement these findings with a crowdsourced user evaluation, where online raters rank model predictions based on alignment with the target user's preference.
We also propose a Large Language Model (LLM)-based personalized rearrangement model that better aligns with real user preferences by explicitly handling objects with multiple valid placement locations, enabling it to accommodate users with flexible placement preferences.


Our work makes the following contributions.
First, we formalize the problem of personalized rearrangement in partially arranged environments.
Second, we introduce PARSEC as an evaluation framework for the above problem. The benchmark features a novel dataset of 110K rearrangement examples collected from 72 real users, covering 93 household objects across 15 environment instances~\footnote{The code and data are available at  \href{https://github.com/kartikvrama/parsec}{https://github.com/kartikvrama/parsec}.}.
Our dataset captures more diverse and flexible preferences than prior rule-based personas, as discussed in Section~\ref{section:dataset}.
Third, we propose ContextSortLM, an LLM-based personalized rearrangement model that explicitly accounts for objects with multiple valid placement locations using a JSON-style preference representation derived from \textit{prior} scene context. Our unique preference representation allows the model to accommodate users with flexible preferences, or lenient object placement constraints, and place objects in a partially arranged environment without disrupting its existing organization.
Lastly, we evaluate ContextSortLM and existing personalized rearrangement approaches on the PARSEC benchmark, complemented by a crowdsourced user evaluation with 108 online raters assessing the alignment of different model predictions to user preferences. 
Our combined results demonstrate the superior performance of models integrating multiple scene context sources for personalized object placement while revealing challenges in modeling environment semantics, which can lead to discrepancies in inferred user preferences. 
Moreover, ContextSortLM outperforms other models in computational evaluations and ranks among the top two in all three environment categories, as rated by online evaluators, emphasizing the advantage of its structured preference representation. We summarize the key takeaways from our evaluation experiments in Section~\ref{section:summary} to guide future work.

\vspace{-0.5em}
\section{Problem Formulation}
\label{section:problem-definition}
We formalize personalized object rearrangement in partially arranged environments as an extension of the rearrangement problem formulation in previous work~\cite{ramachandruni2023consor}. Given an environment with placeable surfaces denoted by the set $\mathcal{S}$, we define an \textit{arrangement} as a set of object-surface pairs representing the contents of each surface: $\mathcal{A} = \{(x, s)\}$, where $x$ denotes an object placed on surface $s$. Object $x$ is not constrained by a fixed closed set and is described using natural language labels. The set of surfaces $\mathcal{S}$ is fixed within an environment but may vary across environments.

In this task, a robot must adapt to a user's object rearrangement preferences by relying on observation instead of explicit user input to then place objects in an environment occupied with objects.
The robot's goal is to place a new set of objects $\mathcal{X}_U$ on surfaces $S$ to transform the environment from its current arrangement $\mathcal{A}_P$ into $\mathcal{A}_G$. The environment may be empty ($\mathcal{A}_P = \Phi$) or partially arranged, meaning that it already contains a few objects. 
{Optionally, the robot has access to prior observations of the environment, denoted as $\mathbf{A}_O = \{\mathcal{A}_O^1, \mathcal{A}_O^2, \dots, \mathcal{A}_O^N\}$, which may include various object configurations.}
The length of observation history $N$ depends on the rearrangement technique; many rearrangement models~\cite{abdo2016organizing, brawner2016learning, kapelyukh2022my, newman2024bootstrapping} are not limited by the number of observations, but some approaches only consider a single observation ($N=1$)~\cite{wu2023tidybot, trabucco2022simple} or do not use any prior observations ({$N=0$})~\cite{ramachandruni2023consor, ramrakhya2024seeing}. Similarly, some rearrangement techniques~\cite{abdo2016organizing, kapelyukh2022my, wu2023tidybot} do not model the environment's current arrangement and assume the environment is empty.

The robot has access to two sources of semantic scene context. Prior scene context, or $\mathbf{A}_O$, is derived from observations of the user arranging objects. Current scene context, or $\mathcal{A}_P$, comes from the placement of objects in the initial state of the environment. Figure~\ref{fig:problem-formulation} illustrates how a robotic agent combines both context sources to place new objects.
The core challenge is attending to specific contextual cues from \textit{prior} and \textit{current} scene context to reason about the placement of each unplaced object. This becomes particularly difficult when users have flexible preferences, or allow multiple valid placements for certain objects. 
For example, the user places the peanut butter jar on different surfaces in Figure~\ref{fig:problem-formulation} (red box), but ultimately prefers to keep it on the bottom-left shelf next to the bread (black dotted box), as they want to group sandwich ingredients and snacks together.
In a similar manner, rearrangement models must reason jointly over both context sources to place objects in ways that reflect real-world organizational habits in partially arranged environments.


\vspace{-0.5em}
\section{Prior Work in Personalized Rearrangement}
In this section, we summarize prior work in personalized rearrangement approaches, categorizing models based on their input modality and reliance on \textit{prior} or \textit{current} scene context. 
We also assess their generalization capabilities to new objects and environments and discuss relevant datasets for our benchmark.

\subsection{Vision and Graph Representations for Personalized Rearrangement}
\label{section:related_work_modalities}
Prior work on inferring user preferences from observed object arrangements falls under one of two categories, based on the modality used to represent object arrangements: \emph{vision-based rearrangement models} that infer object placement directly from visual observations~\cite{ramrakhya2024seeing, trabucco2022simple, newman2024degustabot}, and \emph{graph-based rearrangement models} that determine object placement from abstracted scene graphs of object arrangements~\cite{ramachandruni2023consor, kapelyukh2022my, wu2023tidybot, abdo2016organizing, brawner2016learning, sarch2022tidee}.
The choice of modality influences the precision of object placement and generalization capabilities to new objects and unseen environments.

\textbf{Vision-based rearrangement models} infer object placements directly from RGB or RGB-D observations. These techniques achieve precise object placement by predicting geometrical coordinates, either through a neural vision-to-placement policy~\cite{ramrakhya2024seeing}, a vision-language model (VLM)~\cite{newman2024degustabot} or a search algorithm utilizing semantic-geometric maps~\cite{trabucco2022simple}.
Vision-based approaches also adapt to new user preferences, either through \textit{prior}~\cite{newman2024degustabot, trabucco2022simple} or \textit{current} scene~\cite{ramrakhya2024seeing} semantic context.
However, these methods have some trade-offs. Trabucco et al.~\cite{trabucco2022simple} cannot rearrange objects unobserved in prior observations, and Ramrakhya et al.~\cite{ramrakhya2024seeing} cannot rearrange new objects.
Newman et al.~\cite{newman2024degustabot} generalizes to new objects through VLMs, but is restricted to performing table-top rearrangement and does not generalize to more complex environments. 

\textbf{Graph-based rearrangement approaches} represent object arrangements as scene graphs, where nodes correspond to objects and placeable surfaces, and edges represent object-surface placements. 
These methods infer object placement from scene graph through neural network rearrangement policies~\cite{kapelyukh2022my, ramachandruni2023consor, sarch2022tidee} or by transforming into alternate representations, such as text~\cite{wu2023tidybot, wang2024apricot} or pairwise ranking matrices~\cite{abdo2016organizing, brawner2016learning}.
Graph-based rearrangement models generalize to unseen objects by leveraging semantic similarities among objects, either through explicit knowledge graphs~\cite{abdo2016organizing, brawner2016learning} or by using pre-trained semantic embeddings~\cite{kapelyukh2022my, ramachandruni2023consor}.
However, most graph-based approaches trade off precise placement, typically assigning objects to predefined surfaces rather than outputting exact coordinates.
Moreover, some methods either disregard semantic information from the environment~\cite{ramachandruni2023consor, abdo2016organizing}, or overfit to a single environment~\cite{kapelyukh2022my, brawner2016learning}.
Wu et al.~\cite{wu2023tidybot} and Wang et al.~\cite{wang2024apricot} provide notable exceptions by incorporating LLMs, enabling generalization to new environments.

While vision and graph-based rearrangement models have complementary strengths in adapting to new users, graph-based models excel at generalizing to unseen objects -- an essential capability for robots working in human environments.
Since directly comparing these model types is challenging, we will focus exclusively on graph-based approaches for the remainder of the paper.

\begin{table}[ht]
    \centering
    \resizebox{0.48\textwidth}{!}{%
    \begin{tabular}{|l|l|l|l|l|}
        \hline
         & \multicolumn{2}{|c|}{\textbf{Scene Context}} & \multicolumn{2}{|c|}{\textbf{Models Semantics about X}} \\
        \cline{2-5}
        \textbf{Rearrangement Model} & \textbf{Prior Observations} & \textbf{Current Scene} & \textbf{Environment} & \textbf{Objects} \\
        \hline
        CF \cite{abdo2016organizing} & \ding{51} &  &  & \ding{51} \\
        \hline
        NeatNet \cite{kapelyukh2022my} & \ding{51} &  &  & \ding{51} \\
        \hline
        TidyBot \cite{wu2023tidybot} & \ding{51} &  & \ding{51} & \ding{51} \\
        \hline
        ConSOR \cite{ramachandruni2023consor} &  & \ding{51} & & \ding{51} \\
        \hline
        CF+ \cite{brawner2016learning} & \ding{51} & \ding{51} & & \\
        \hline
        APRICOT \cite{wang2024apricot} & \ding{51} & \ding{51} & \ding{51} & \ding{51} \\       
        \hline
        \textbf{ContextSortLM (Ours)} & \ding{51} & \ding{51} & \ding{51} & \ding{51} \\       
        \hline
    \end{tabular}
    }
    \caption{\small{Comparison of existing rearrangement approaches that adapt to new user preferences, categorized by their use of semantic context for preference adaptation and the semantic information encoded by the rearrangement model. Our proposed model, ContextSortLM, integrate scene context from prior and current observations via a unique preference representation to ensure user-aligned object placements.}}
    \label{tab:rearrangement}
\end{table}
Table~\ref{tab:rearrangement} compares existing graph-based rearrangement methods based on their use of semantic context for preference adaptation
and the semantic information encoded by each model.
Few rearrangement models use both \textit{prior} and \textit{current} scene context for preference adaptation~\cite{brawner2016learning, wang2024apricot, sarch2022tidee}. However, Brawner and Littman~\cite{brawner2016learning} fail to handle unseen objects and new environments, and Wang et al.~\cite{wang2024apricot}'s model relies on user interaction to disambiguate user preferences and is designed specifically for a fridge organization task.
Sarch et al.~\cite{sarch2022tidee} also leverages \textit{prior} and \textit{current} scene context to determine plausible object-receptacle pairs, but their model memorizes a single user's preferences during training and cannot adapt to new user preferences without retraining.
Our proposed approach, ContextSortLM, adapts to new user preferences by integrating \textit{prior} and \textit{current} scene context to place objects in partially arranged environments without user supervision. ContextSortLM also leverages LLM reasoning to generalize to unseen objects and new environments. We present the details of this model in Section~\ref{section:methods}.

\subsection{Personalized Rearrangement Datasets}

Existing datasets for personalized object rearrangement include rule-based~\cite{ramachandruni2023consor, wu2023tidybot} and user-generated~\cite{ kapelyukh2022my, newman2024degustabot} datasets of object arrangements.
Rule-based datasets~\cite{ramachandruni2023consor, wu2023tidybot} are generated from pre-defined organizational rules that dictate object placement and grouping, such as `organize objects by their affordance'~\cite{ramachandruni2023consor} or `put shirts on the sofa and other clothes in the closet'~\cite{wu2023tidybot}. These rule-based datasets scale well, but lack the nuances and diversity of real user preferences, as we demonstrate in Section~\ref{section:dataset}.
Alternately, user-generated datasets~\cite{kapelyukh2022my, newman2024degustabot} collected by online workers performing grounded organizational tasks~\cite{newman2024bootstrapping, toris2015unsupervised, kapelyukh2022my}, capture diverse placement preferences and record fine-grained placements.
However, the data is collected for very specific organizational tasks in 1-2 fixed environments and contains a limited number of object categories and environments.
Addressing the limitations of prior datasets, we collect a crowdsourced dataset of 432 object arrangements from real users that spans 5 different organizational tasks and involves various object categories and environments.

\vspace{-0.5em}
\section{PARSEC Benchmark and Dataset}
\label{section:dataset}

For the PARSEC benchmark, we seek a dataset that a)~incorporates data from real people instead of rules, b)~includes multiple examples for each user, and c)~spans a diverse set of object categories and environments. Since existing datasets do not meet our criteria, we collected our own data by hiring online workers to arrange household objects in environments resembling real homes~\footnote{Data collection was IRB exempt. Workers were hired from Prolific.}. 
Each online worker separately arranged six sets of objects in a single environment accompanied with short descriptions. The object sets were sampled to include objects relevant to the environment type and random household objects, and workers were instructed to only arrange the objects they considered relevant.

\begin{figure}[t]
    \centering
    \includegraphics[width=0.43\textwidth]{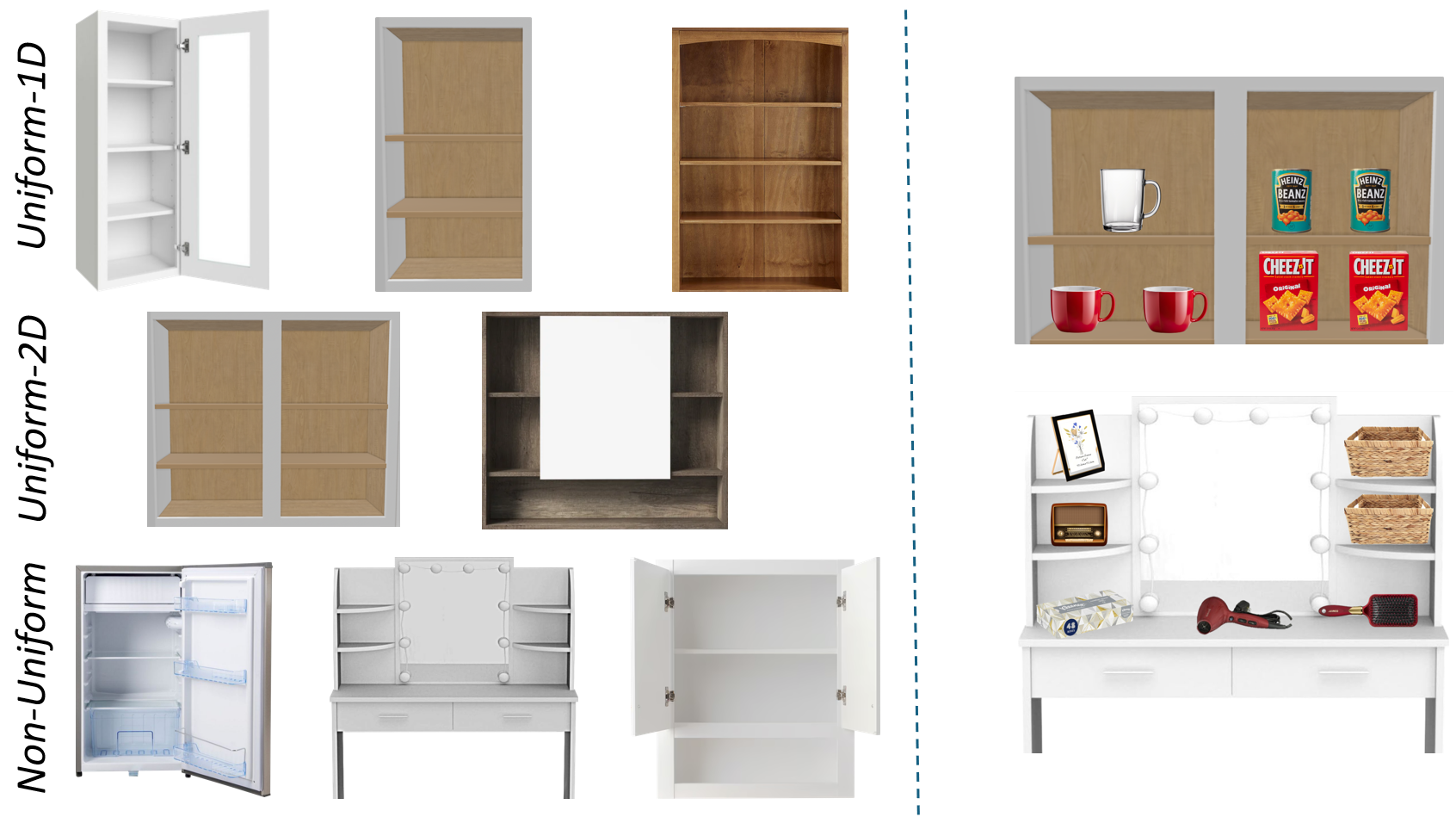}
    \caption{\small{The environments in the PARSEC benchmark can be categorized by the number of surface types and their position. A and B illustrate some examples of real user arrangements from the dataset.}}
    \label{fig:dataset}
    \vspace{-2em}
\end{figure}
In total, we collected 432 object arrangements, involving 93 household objects and spanning 72 users~\footnote{Out of 75 users, three were removed due to failed attention checks.} and 15 environment instances. We uniformly sampled the environments from five household organizational tasks: stocking a kitchen pantry, arranging a bathroom cabinet, rearranging a bedroom dresser, stocking a fridge, and decorating a display shelf. These environments fall under three semantic categories, as shown in Figure~\ref{fig:dataset}:
Environments with identical surface types and surfaces positioned vertically, or \code{Uniform-1D}; environments with identical surface types, arranged in a 2D configuration, or \code{Uniform-2D}; and environments with more than one surface type, or \code{Non-Uniform}.

Each category presents unique challenges for adapting to user preferences.
In environments with multiple identical surfaces (\code{Uniform-1D} and \code{Uniform-2D}), users pay more attention to which objects to group than the exact surface to place them. For instance, in example A of Figure~\ref{fig:dataset}, a user may prioritize grouping all types of mugs but place them in the left or right shelves interchangeably.
In contrast, environments with multiple surface types (\code{Non-Uniform}) encourage users to assign objects to specific surfaces. However, these preferences become more nuanced when multiple instances of each surface type exist. For example, in example B, users assigns objects to separate surfaces (e.g., self-care items on the table) while maintaining specific organizational patterns among identical surfaces (e.g., mementos on the left shelf and baskets on the right). 
Modeling the relative positions of surfaces in two dimensions (\code{Uniform-2D} and \code{Non-Uniform}) adds further complexity when generalizing to unseen environments.

\smallskip
\noindent \textbf{Dataset Generation}: From the crowdsourced object arrangements, we created a dataset of 110K examples.
Each user annotator $m$ provided six object arrangements within the same environment instance. Each object arrangement $\mathcal{A}_i = \{(x, s)\}$ comprises objects $x$, represented by text labels of its semantic category, and surfaces $s$, described as a tuple of surface type and relative 2D position. Optionally, $s$ can be expressed as a templated language description, such as `top-right shelf'.
To create rearrangement examples, we iteratively selected one arrangement as the target arrangement 
$\mathcal{A}_G^* = \mathcal{A}_i$
and designated the other five as observed user arrangements
$\mathbf{A}_O=\{\mathcal{A}_j | j\neq i\}$,
generating $\binom{5}{2}$ pairs of
($\mathbf{A}_O$, $\mathcal{A}_G^*$).
For each target arrangement $\mathcal{A}_G^*$, we created several pairs of partially arranged states $\mathcal{A}_P$ and unplaced object sets $\mathcal{X}_U$ by randomly omitting objects from $\mathcal{A}_G^*$, labeling the omitted objects as the unplaced set.
Empty environment states were also constructed by removing all objects from $\mathcal{A}_P$.
The resulting dataset is represented as a set of tuples: $\mathcal{D} = \{(m, \mathbf{A}_O, \mathcal{A}_P, \mathcal{X}_U)\}$.
Given $\mathbf{A}_O$ and $\mathcal{A}_P$, the rearrangement model must place $\mathcal{X}_U$ alongside $\mathcal{A}_P$ while adhering to $m$'s preferences.  

\begin{figure}[t]
    \centering
    \includegraphics[width=0.3\textwidth]{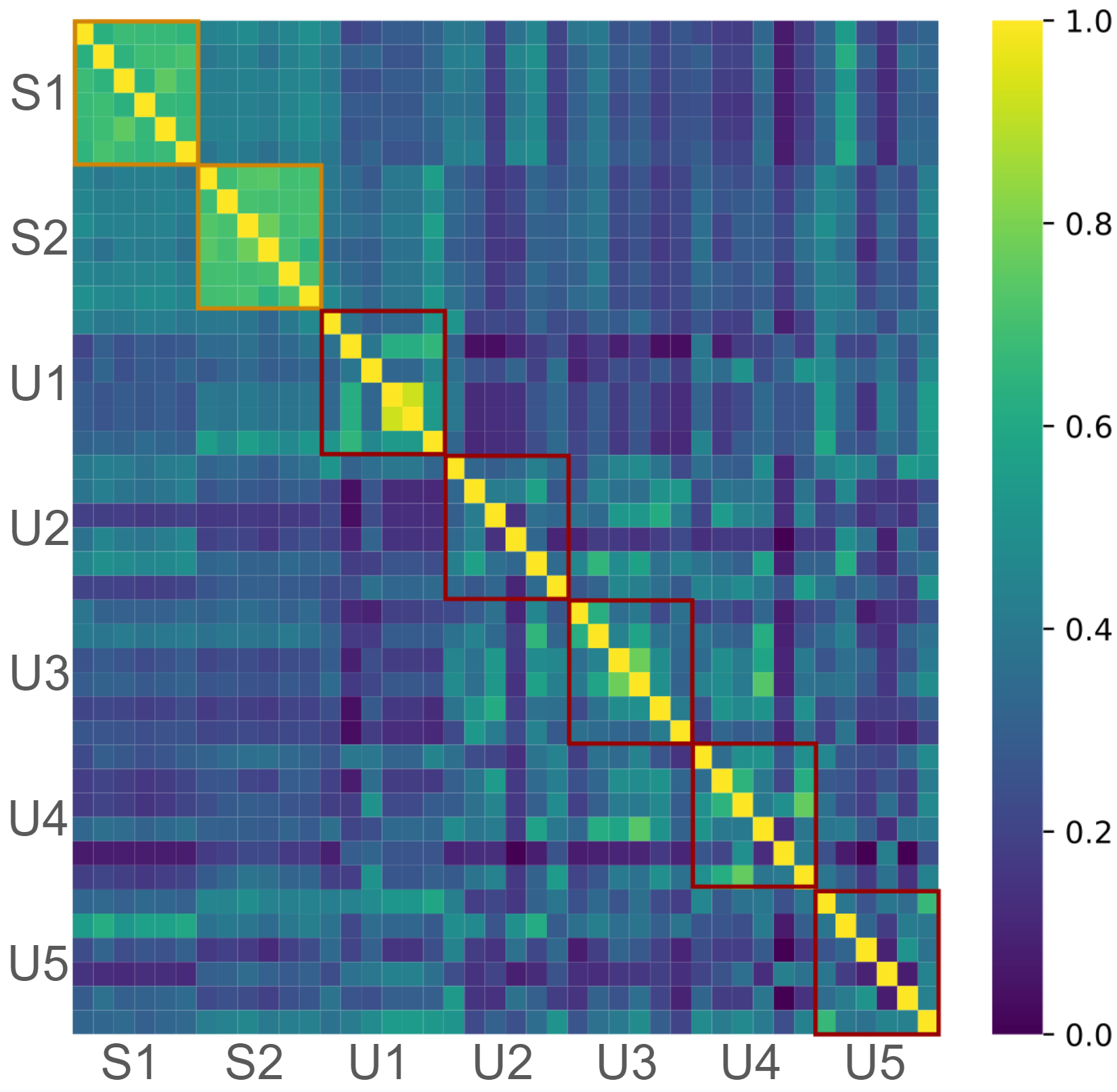}
    \caption{\small{To demonstrate the within-user variability and across-user diversity of crowdsourced rearrangement data, we plot the average WordNet similarity among both rule-based (S1, S2) and real user object arrangements (U1-U5) of a fridge environment.}}
    \label{fig:pairwise-comparison}
    \vspace{-1.4em}
\end{figure}

\smallskip
\noindent \textbf{Comparing Manually Defined and Real User Preferences}: We compared our crowdsourced data from a fridge in the \code{Non-Uniform} category  with two rule-based user personas resembling the sorting criteria from \cite{wu2023tidybot}, with preference rules such as `Put soda cans and eggs on the top door shelf' and `Put produce on the middle shelf'. Figure~\ref{fig:pairwise-comparison} visualizes the average WordNet similarity between rule-based arrangements (S1, S2) and real user arrangements (U1–U5). 
We find that real user arrangements exhibit lower within-user similarity scores ($Si,Si$) compared to rule-based arrangements ($Uj, Uj$) and show higher variance in between-user similarity scores ($Si,Sj$) than rule-based arrangements ($Ui,Uj$), indicating that object arrangements in PARSEC are more lenient and diverse than rule-based personas. The diversity and leniency of real-world placement preferences makes adaptation more challenging, motivating our decision to collect crowdsourced data.

\begin{SCfigure*}
    \centering
    \includegraphics[width=0.65\textwidth]{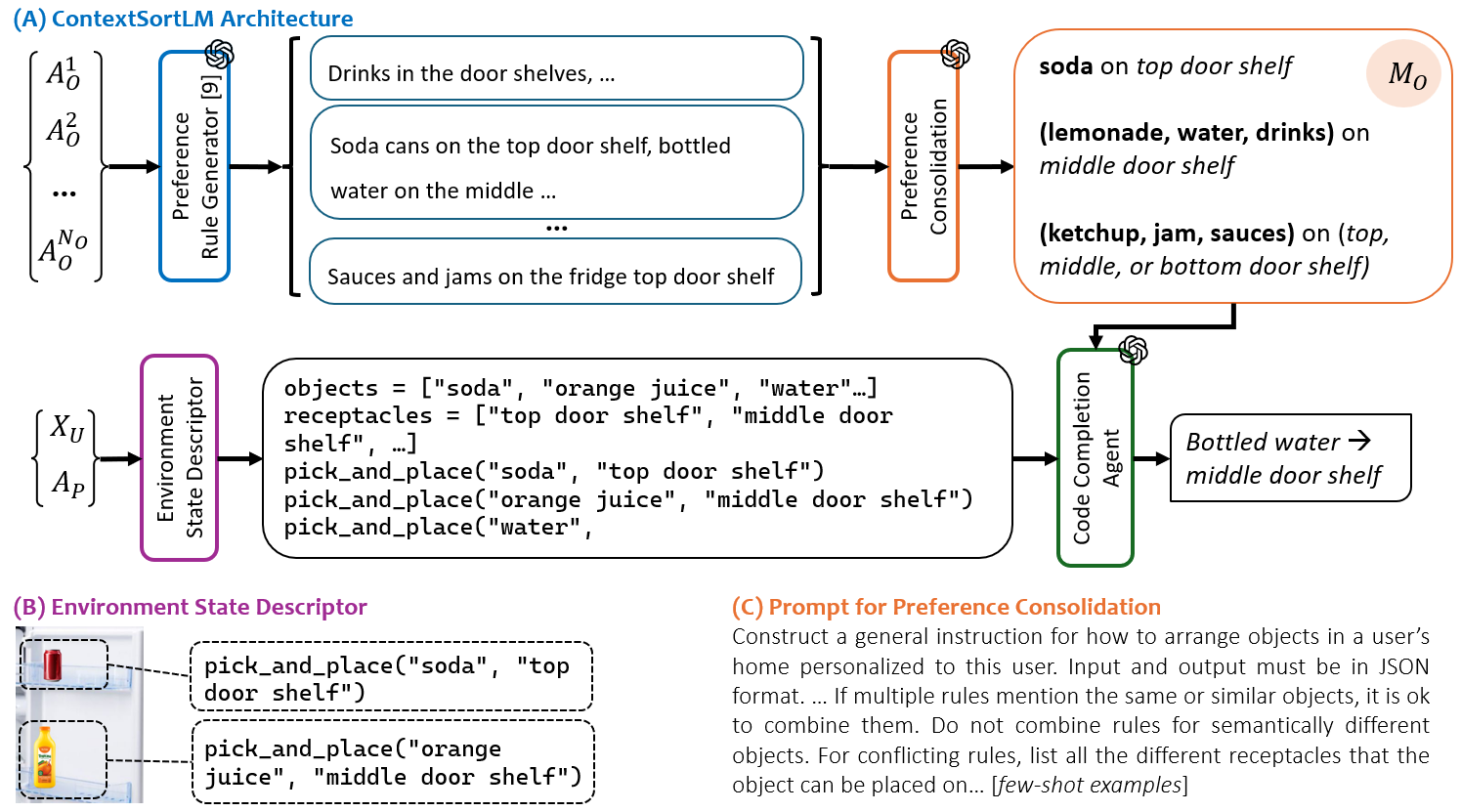}
    \caption{\small{Model architecture illustrating how ContextSortLM consolidates observed arrangements into a structured preference representation that captures multiple valid object placements and uses it to place objects in partially arranged environments.}}
    \label{fig:contextsortlm}
    \vspace{-1.2em}
\end{SCfigure*}

\vspace{-0.5em}
\section{Algorithm Selection}
\label{section:methods}
We include all the graph-based rearrangement algorithms from Table~\ref{tab:rearrangement} in our evaluation. 
Of these algorithms, the CF~\cite{abdo2016organizing}, NeatNet~\cite{kapelyukh2022my}, and TidyBot~\cite{wu2023tidybot} techniques infer user preferences from \textit{prior} scene context, and ConSOR adapts to preferences from \textit{current} scene context.
TidyBot-Random is a variant of TidyBot that samples 
a random arrangement $\mathcal{A}_O^i$ from observed arrangements $\mathbf{A}_O$ to generate preference rules, since TidyBot requires a single example.
In contrast, the CF+~\cite{brawner2016learning} and APRICOT~\cite{wang2024apricot} models combine \textit{prior} and \textit{current} scene context to place objects in partially arranged environments.
APRICOT-NonInteractive, adapted from the `Non-Interactive' baseline in the work by Wang et al.~\cite{wang2024apricot}, executes personalized object placement by constructing a single textual preference description from $\mathbf{A}_O$.

\smallskip
\noindent \textbf{ContextSortLM}: 
To better align with flexible user preferences, we propose ContextSortLM, a personalized rearrangement model that handles multiple valid object placements in partially arranged environments by leveraging LLM summarization and code completion capabilities, as shown in Figure~\ref{fig:contextsortlm} (A).
Existing LLM-based rearrangement methods~\cite{wu2023tidybot, wang2024apricot} model user preferences as textual descriptions of observed arrangements.
However, they fail to capture flexible preferences, as they tend to either overfit to a single arrangement~\cite{wu2023tidybot} or overgeneralize~\cite{wang2024apricot}, since LLMs often oversimplify information when summarizing~\cite{birhane2023science}.
To address this, ContextSortLM generates a JSON-style preference representation $M_O$ from observed arrangements $\mathbf{A}_O$ that prioritizes object grouping and accommodates multiple valid placements. Our model then reasons jointly over the preference representation $M_O$ and the environment's current arrangement $\mathcal{A}_P$ to better align with flexible user preferences when placing items in partially arranged environments.
ContextSortLM first extracts preference rules for each object arrangement in $\mathbf{A}_O$ using an LLM rule-generation agent inspired by prior work~\cite{wu2023tidybot}. 
Rather than simply joining these rules -- which can produce conflicting preferences for the same object category -- our model consolidates them into a single, consistent representation $M_O$ through a separate LLM call (Figure~\ref{fig:contextsortlm} (C)) that explicitly accounts for multiple valid object placements.
Objects are then placed by asking the LLM to complete a code completion prompt, with $\mathcal{A}_P$ rewritten as Python-style `pick-place' commands -- as shown in Figure~\ref{fig:contextsortlm} (B) -- and $M_O$ appended to it.

\begin{table*}[t]
\centering
\resizebox{0.72\textwidth}{!}{%
\begin{tabular}{lcccc|cccc}
\hline
\multicolumn{1}{c}{\textbf{Model}}         & \multicolumn{4}{c}{\textbf{{KnownEnv}}} & \multicolumn{4}{c}{\textbf{{NovelEnvCategory}}} \\
\hline
                       & \code{Uniform-1D} & \code{Uniform-2D} & \code{Non-Uniform} & \textbf{Average} 
                       & \code{Uniform-1D} & \code{Uniform-2D} & \code{Non-Uniform} & \textbf{Average} \\
\hline
ContextSortLM (Ours)~$\dagger$  & \textbf{0.54} & \textbf{0.57} & \textbf{0.65} & \textbf{0.59} & \textbf{0.54} & \textbf{0.57} & \textbf{0.65} & \textbf{0.59}\\
APRICOT-NonInteractive~$\dagger$ & 0.50 & 0.50 & 0.56 & 0.53 & 0.50 & 0.50 & 0.56 & 0.53 \\
TidyBot-Random~$\dagger$   & 0.46 & 0.40 & 0.46 & 0.44 & 0.46 & 0.40 & 0.46 & 0.44\\
ConSOR           & 0.36 & 0.41 & 0.35 & 0.37 & 0.32 & 0.38 & 0.27 & 0.31\\
CF               & 0.30 & 0.33 & 0.23 & 0.28 & 0.30 & 0.31 & 0.24 & 0.28\\
CFFM             & 0.23 & 0.22 & 0.23 & 0.23 & -    & -    & -    & -\\
NeatNet          & 0.28 & 0.25 & 0.29 & 0.28 & -    & -    & -    & -\\
\hline
\end{tabular}%
}
\caption{\small{Placement Accuracy ($PA$) calculated when adapting to new users in seen environments and unseen-category environments. The $\dagger$ indicates models not trained on any example from the dataset, meaning that they perform identically in KnownEnv and NovelEnvCategory conditions.}}
\label{tab:placement_accuracy_envs}
\vspace{-1.4em}
\end{table*}

\vspace{-0.5em}
\section{Evaluation on PARSEC Scenarios}
\label{section:benchmark-evaluation}

We utilize k-fold cross-validation to evaluate the rearrangement methods presented in Table~\ref{tab:rearrangement}. Each fold in the cross-validation set includes a training set $\mathcal{D}_{train}$ and test set $\mathcal{D}_{test}$. We conduct two experiments -- the KnownEnv and NovelEnvCategory experiments -- to evaluate adaptation to user preferences in previously seen and unseen environments. We generate each fold by excluding the examples from one of five users per environment category and examples of all users from one of three environment categories from training respectively. On average, each fold in KnownEnv contains 2806 $\mathcal{D}_{train}$ and 701 $\mathcal{D}_{test}$ examples, and each fold in NovelEnvCategory contains 2338 $\mathcal{D}_{train}$ and 1169 $\mathcal{D}_{test}$ examples. 
In each fold, we separate arrangement examples of a random user from $\mathcal{D}_{train}$ as a validation set $\mathcal{D}_{val}$, resulting in approximately 85 $\mathcal{D}_{val}$ examples per fold. We applied early stopping during model training based on the average number of misplaced objects calculated on $\mathcal{D}_{val}$ examples.

CF+ and NeatNet are trained and evaluated on the same environment instance in KnownEnv and do not generalize to new environments. In contrast, TidyBot-Random, APRICOT-NonInteractive, and ContextSortLM are neither trained nor provided examples from this dataset and are always evaluated on unseen users and environments~\footnote{ContextSortLM, APRICOT-NonInteractive, and TidyBot-Random use the \textit{gpt-4-0613} model.}.

\begin{figure}[t]
    \centering
    \includegraphics[width=0.3\textwidth]{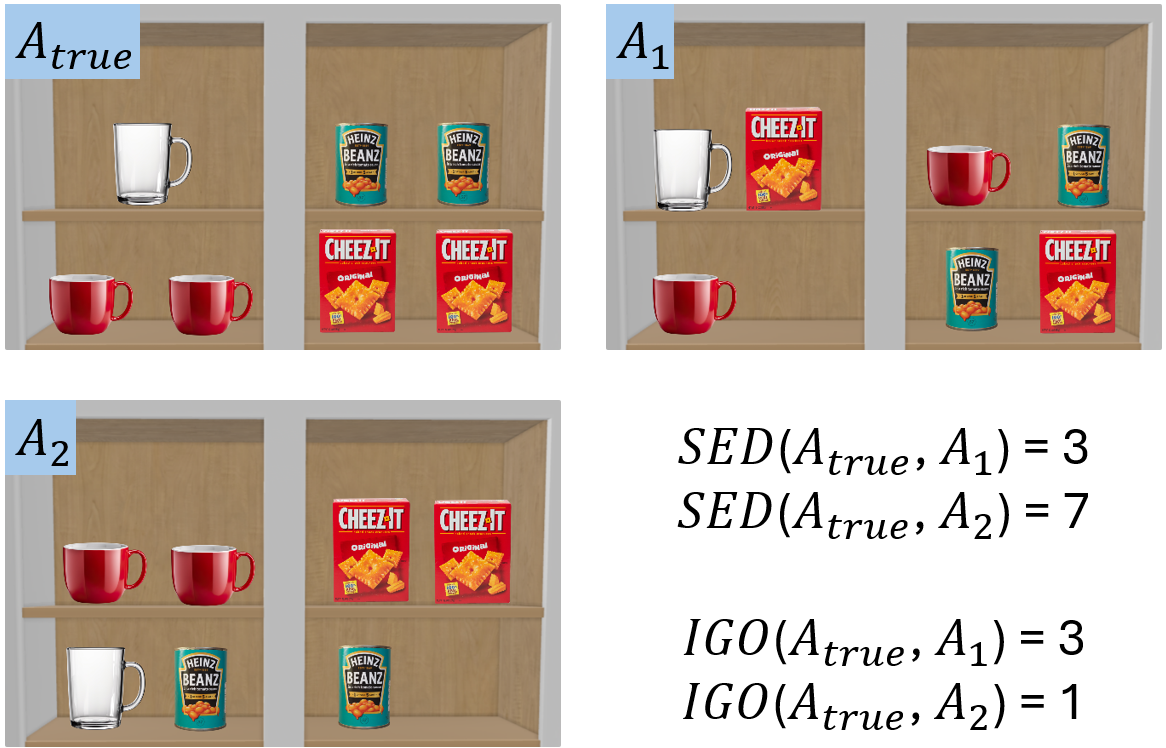}
    \caption{\small{$SED$ and $IGO$ calculated between a hypothetical user arrangement $A_{true}$ and two possible predicted arrangements $\mathcal{A}_a$ and $\mathcal{A}_2$. Note how $\mathcal{A}_2$ has a high $SED$ but a low $IGO$, since most same-category objects are grouped together as in $\mathcal{A}_{true}$ but not placed on the correct surface.}}
    \label{fig:metrics}
    \vspace{-2em}
\end{figure}

\smallskip
\noindent \textbf{Metrics}: We define two measures of similarity between the model's predicted arrangement $\mathcal{A}_G$ and the user's true arrangement $\mathcal{A}_{G}^*$ to assess model performance, shown in Figure~\ref{fig:metrics}.
The \textit{Scene Edit Distance}, or $\mathbf{SED}$, borrowed from prior work~\cite{ramachandruni2023consor}, is the minimum number of objects that must be moved in $\mathcal{A}_G$ to perfectly match $\mathcal{A}_{G}^*$.
The \textit{Number of Incorrectly Grouped Objects}, or $\mathbf{IGO}$, is the minimum number of objects that must be moved in $\mathcal{A}_G$ so that the same sets of objects are placed together as in $\mathcal{A}_{G}^*$, while ignoring the exact surface on which the objects are placed.
Collectively, the $SED$ and $IGO$ measure deviation from the preferred  surface assignment and object grouping respectively.
We also compute the Placement Accuracy or $PA$, which is the average number of object placements predicted for $\mathcal{X}_U$ that match the placements in $\mathcal{A}_{G}^*$.

\label{section:metrics}


\subsection{Results}
\label{section:results-aggregate}
We present the results of KnownEnv and NovelEnvCategory experiments. We report $PA$ as an aggregate measure of rearrangement performance for each experiment. For the KnownEnv, we also report $SED$ and $IGO$ scores across different initial environment conditions to study how rearrangement performance changes over increasing \textit{current} scene context.

\smallskip
\noindent \textbf{Aggregate Performance}: Table~\ref{tab:placement_accuracy_envs} presents the $PA$ scores for KnownEnv and NovelEnvCategory experiments.
Across both experiments, ContextSortLM and APRICOT-NonInteractive achieve higher $PA$ than TidyBot-Random and ConSOR, demonstrating the benefit of integrating multiple semantic context sources rather than relying on a single source for preference adaptation. 
Moreover, ContextSortLM, APRICOT-NonInteractive, and TidyBot-Random have a higher $PA$ than other methods without any pre-training or using in-context examples from our dataset, highlighting the importance of pre-trained commonsense reasoning in adapting to new user preferences. 

ContextSortLM outperforms APRICOT-NonInteractive in all environment categories, emphasizing the benefit of its structured preference representation over a textual description.
Notably, the $PA$ of ContextSortLM and APRICOT-NonInteractive differs most in the \code{Non-Uniform} category, where more surfaces typically lead to fewer objects per surface.
This likely occurs because APRICOT-NonInteractive’s LLM-generated textual preference descriptions overgeneralize user preferences, causing objects to be over-grouped on surfaces.

Among non-LLM models, ConSOR considerably outperforms the bottom half of the table (CF, CFFM, and NeatNet). NeatNet and CFFM struggle due to limited training data per environment, but CF underperforms even after training across environments. CF's reliance on pairwise object similarity misses broader contextual cues essential for rearranging across different environments, resulting in lower placement accuracy.
CFFM's performance also suffers due to both limited data and a lack of global semantic context, despite using both scene context sources. Incorporating global semantic context is, therefore, essential for modeling preferences across environments.

CF's $PA$ remains relatively stable across KnownEnv and NovelEnvCategory experiments compared to ConSOR's $PA$, particularly in the \code{Non-Uniform} category, suggesting that models leveraging \textit{prior} scene context generalize better to unseen environments than those relying on \textit{current} scene context. However, more experiments are required.

\begin{SCfigure*}
    \centering
    \includegraphics[width=0.7\textwidth]{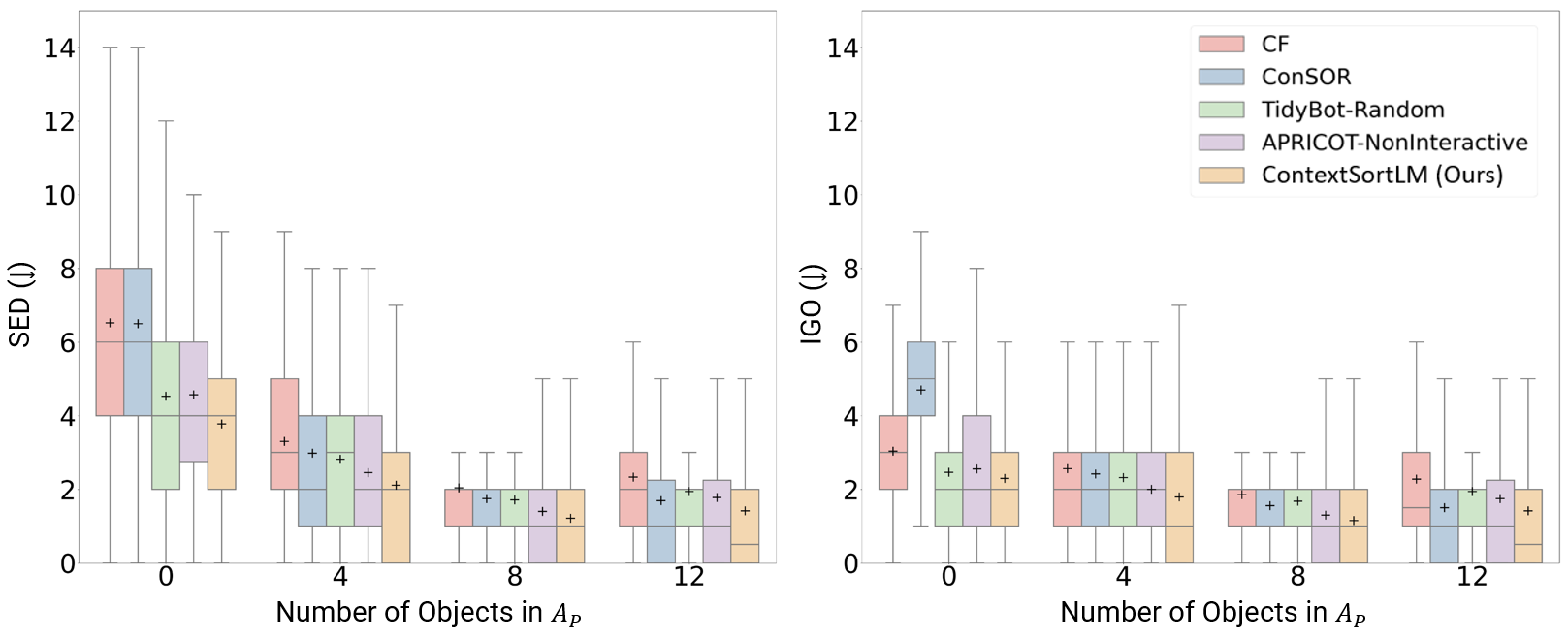}
    \caption{\small{Placement error metrics ($SED$ and $IGO$) calculated as a function of the number of objects in the environment's current arrangement for the KnownEnv experiment. The `+' sign denotes the mean value.}}
    \label{fig:new-user-known-env}
    \vspace{-1.4em}
\end{SCfigure*}

\smallskip
\noindent \textbf{Placement Error in Partially Arranged Environments}: Figure~\ref{fig:new-user-known-env} shows the placement error ($SED$ and $IGO$ scores) of the above models as a function of $N_P$, the number of objects in the environment's current arrangement $\mathcal{A}_P$. $N_P=0$ represents an empty environment and $N_P=12$ a densely populated one.
ContextSortLM achieves the lowest mean $SED$ in both sparsely occupied ($N_P = \{0, 4\}$) and densely occupied environments ($N_P = \{8, 12\}$), with APRICOT-NonInteractive ranking second, which further supports integrating multiple scene context sources for personalized rearrangement. 
Moreover, ConSOR and TidyBot-Random have comparable mean $SED$ scores in densely occupied environments, indicating that \textit{current} scene context is as effective as \textit{prior} scene context for object placement in partially arranged environments.
Despite achieving a lower mean $SED$ overall, ContextSortLM's $SED$ scores exhibit high variance in densely occupied environments, indicating that it struggles to place objects in such scenarios.

Across all models, the $IGO$ scores are lower than $SED$ scores in sparsely occupied $\mathcal{A}_P$, suggesting that rearrangement models are better at grouping similar objects than selecting appropriate surfaces. This is likely because models leverage common object similarities encoded in external knowledge.
Our finding also aligns with human tendencies to agree strongly on object similarity~\cite{mur2013human}.
The $SED$ and $IGO$ scores converge in densely occupied environments ($N_P \geq 8$), where limited empty surfaces make object placement synonymous with grouping similar items.

\vspace{-0.5em}
\section{Evaluation with Online Raters}
\label{section:real-user-evaluation}
To assess the alignment of computational metrics with human judgment, we conducted a crowdsourced user evaluation where online raters ranked different rearrangement models based on alignment with target user's preferences.
We selected four top-performing models -- ContextSortLM, APRICOT-NonInteractive, TidyBot-Random, and ConSOR -- and chose examples from PARSEC where all four model predictions differed.
Online raters in our experiment first examined the observed arrangements, $\mathbf{A}_O$, and wrote a summary describing them. This summary serves as a quality check, allowing us to filter out raters who submit irrelevant summaries.
Raters then reviewed the environment's current arrangement, objects to be placed, and the predicted object arrangements from all four models. Raters identified which predicted object arrangement perfectly matched the target user's preferences and ranked the arrangements based on alignment to user preferences.
To reduce bias, we counterbalanced the order of model predictions and recruited three independent raters per example.
In total, we hired 108 raters~\footnote{Out of 110 raters, two were removed due to poor quality summaries.} to evaluate 36 object arrangement examples, spanning 14 user preferences, each with 2–3 variations of $\mathcal{A}_P$.

\smallskip
\noindent \textbf{Evaluation Metrics}: We define two metrics to analyze rater responses for each rearrangement model.
Alignment Score, or $s_{align}$, is the percentage of raters who found the model's predicted object arrangement to perfectly align with the target user's preferences.  
Rank Score, or $s_{rank}$, is the average rank raters give to a model's predictions. 
A higher $s_{align}$ indicates that the model frequently places objects to match the target user's preferences, and a lower $s_{rank}$ signifies better alignment with the user's preferences compared to other models.

\vspace{-0.2em}
\subsection{Results}
\vspace{-0.2em}

\smallskip
\noindent \textbf{Alignment Score}: Table~\ref{tab:alignment-score} presents the $s_{align}$ metric, which measures how well models match the target user's preferences. 

Our results indicate that models incorporating \textit{prior} and \textit{current} scene context, ContextSortLM and APRICOT-NonInteractive, achieve higher alignment scores than TidyBot-Random and ConSOR across all three environment categories, further validating the use of multiple sources of scene context for preference adaptation. APRICOT-NonInteractive outperforms ContextSortLM in \code{Uniform-1D} and \code{Uniform-2D} categories but underperforms in the \code{Non-Uniform} category, likely because APRICOT-NonInteractive tends to over-cluster similar objects. ConSOR scores higher than TidyBot-Random in the category \code{Uniform-1D} but lags in the \code{Uniform-2D} and \code{Non-Uniform} categories, possibly because \code{Uniform-1D} examples are densely occupied and offer richer semantic context benefiting ConSOR.

\begin{table}[t]
    \centering
    \resizebox{0.42\textwidth}{!}{%
    \begin{tabular}{lccc}
        \toprule
        Rater Response & \code{Uniform-1D} (\%) & \code{Uniform-2D} (\%) &  \code{Non-Uniform} (\%) \\
        \midrule
        ContextSortLM (Ours)    & 40.5 & 37.8  & \textbf{60.7} \\
        APRICOT-NonInteractive   & \textbf{45.2} & \textbf{43.2}  & 7.1  \\
        TidyBot-Random           & 21.4 & 27.0  & 25.0 \\
        ConSOR                   & 26.2 & 24.3  & 28.6 \\
        None                    & 16.7 & 27.0  & 3.6  \\
        \bottomrule
    \end{tabular}%
    }
    \caption{\small{Alignment scores ($s_{align}$), measuring how often each models matches the target user's preference. `None' corresponds to the rater finding none of the model predictions aligning with the target user's preferences.}}
    \label{tab:alignment-score}
    \vspace{-1.4em}
\end{table}

Many raters found no model perfectly aligned with user preferences, especially in Uniform-2D examples, where $27.0\%$ chose `None.'
\code{Uniform-2D} environments feature identical surfaces in a 2D layout, which is challenging to represent semantically, leading to preference mismatches and misplaced objects. This underscores the challenge of modeling spatial information about previously unseen environments such as relative surface positions.
Notably, some raters who selected `None' explicitly mentioned in their summaries that the target user’s preferences were unclear, which may have hindered their ability to identify a perfect match. 

\begin{figure}[t]
    \centering
    \includegraphics[width=0.45\textwidth]{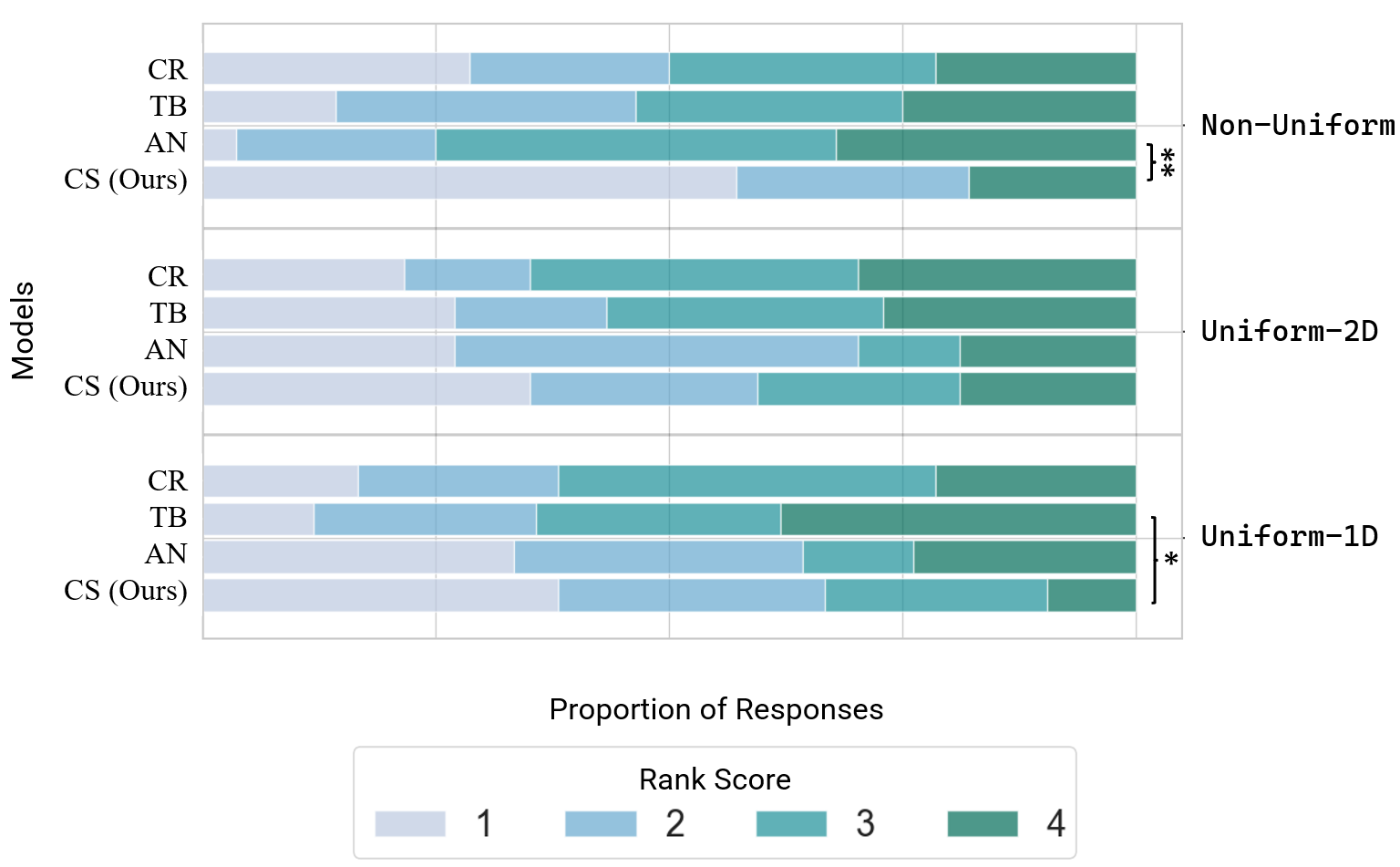}
    \caption{\small{Distribution of rank scores $s_{rank}$, derived from rater-assigned model rankings, and categorized by environment type. `0.5' on the x axis denotes the median rating. The acronyms CR, TB, AN, and CS (Ours) represent ConSOR, TidyBot-Random, APRICOT-NonInteractive, and ContextSortLM, respectively. `*' and `**' symbols indicate p-values less than 0.05 and 0.01 respectively.}}
    \label{fig:user-model-rank}
    \vspace{-1.4em}
\end{figure}

\smallskip
\noindent \textbf{Rank Score}:
Figure~\ref{fig:user-model-rank} presents the $s_{rank}$ metric, derived from rater-assigned model rankings and categorized by environment types. 
We use Friedman's one-way test followed by post-hoc Wilcoxon Signed-Rank tests~\footnote{Bonferroni correction of $\alpha=6$ was applied.} for statical analysis, marked in the figure. Friedman's test indicates statistically significant differences in the \code{Uniform-1D} and \code{Non-Uniform} category ($p < 0.001$), and post-hoc pairwise comparisons with Wilcoxon Signed-Rank test reveal that ContextSortLM has a significantly lower median rank than TidyBot-Random in the \code{Uniform-1D} category ($p < 0.01$) and APRICOT-NonInteractive in the \code{Non-Uniform} category ($p < 0.05$), which is consistent with previous findings. 

Surprisingly, there are few statistically significant differences among models for the \code{Uniform-1D} and \code{Non-Uniform} categories, suggesting that users tolerate reasonable variations in object placement.
The absence of any significant differences in the $\code{Uniform-2D}$ category is likely due to the challenges of accurately modeling spatial information in these environments, resulting in discrepancies in user preferences.

\vspace{-0.5em}
\section{Summary and Discussion}
\label{section:summary}
Our evaluation results strongly support integrating \textit{prior} and \textit{current} scene context for personalized rearrangement. To guide future work in better integrating the two scene context sources, we summarize key takeaways from evaluations and highlight ContextSortLM's limitations.

Among the models integrating dual context sources, ContextSortLM outperforms APRICOT-NonInteractive when comparing computational metrics, particularly in \code{Non-Uniform} environments, and achieves consistently high alignment and low rank scores across environment categories, highlighting the benefit of accounting for multiple valid object placements when representing user preferences.

While $s_{align}$ scores align with trends seen in computational metrics, $s_{rank}$ scores show few significant differences among models, highlighting challenges in encoding semantic information about the environment. Raters tolerated most variations in object placements but penalized irrelevant object placements in designated easy-access surfaces, such as the cabinet's bottom shelf, or purpose-specific locations, such as the fridge's top shelf or vegetable drawer. Encoding more semantic information from the environment- such as better spatial information about surfaces and knowledge of object usage patterns- will improve preference adaptation.

\smallskip
\noindent \textbf{Limitations of ContextSortLM} ContextSortLM struggles in environments occupied with many objects due to over-reliance on its meta-preference $M_O$, sometimes grouping dissimilar objects. ConSOR's success in the same setting suggests a hybrid LLM/specialized-policy approach -- filtering \textit{prior} scene context with LLMs and resolving conflicts with the current environment via learned policies leveraging \textit{current} scene context. Moreover, ContextSortLM's meta preference $M_O$ is sensitive to noisy object placements in the observed arrangements (e.g., a coffee mug randomly placed in the fridge), and we aim to refine ContextSortLM to ignore such outlier placements in future work.

\section{Conclusion}

In conclusion, we introduced PARSEC, an object rearrangement benchmark where robots adapt to user organizational preferences from scene context for object placement in partially arranged environments. PARSEC includes a novel crowdsourced dataset of 110K evaluation examples collected from 72 real users, covering 93 household objects across 15 environment instances. 
To better align with real-world organizational habits, we proposed ContextSortLM, an LLM-based personalized rearrangement model that accommodates flexible user preferences by explicitly accounting for multiple valid object placements when placing items in partially arranged environments.
We evaluated ContextSortLM and existing personalized rearrangement models on PARSEC and complemented these findings with a crowdsourced user evaluation of 108 online raters ranking model predictions based on alignment to user preferences.
Our results highlight the importance of integrating multiple sources of scene context for personalized object placement in partially arranged environments. However, there are challenges in modeling environment semantics -- such as the environment's spatial layout and utility of different environment surfaces -- leading to discrepancies in inferred user preferences. 
Moreover, ContextSortLM outperforms other models in computational evaluations and ranks among the top two in all three environment categories according to online evaluators. Despite this, ContextSortLM struggles in densely occupied environments, highlighting the need for better techniques to integrate prior and within-scene context.

Finally, we note that the choice of using an LLM for our proposed approach is based on the strong performance of prior LLM-based object rearrangement models~\cite{wu2023tidybot, wang2024apricot, newman2024degustabot}.
As LLM performance is highly dependent on the information provided in the prompt, our work provides design guidelines for future LLM-based rearrangement models. 

\vspace{-0.3em}
\setstretch{0.85}
\bibliographystyle{ieeetr}
\bibliography{references}
\end{document}